\documentclass{article}

\usepackage{microtype}
\usepackage{graphicx}
\usepackage{subfigure}
\usepackage{booktabs} % for professional tables

\usepackage{hyperref}

\usepackage[accepted]{preprint}

\usepackage{amsmath}
\usepackage{amssymb}

\def\Exp{\mathop{\mathbb{E}}}

\icmltitlerunning{End-to-end Generative ZSL via FSL}

\begin{document}

\twocolumn[
\icmltitle{End-to-end Generative Zero-shot Learning via Few-shot Learning}




\begin{icmlauthorlist}
\icmlauthor{Georgios Chochlakis}{ntua}
\icmlauthor{Efthymios Georgiou}{ntua}
\icmlauthor{Alexandros Potamianos}{ntua,bs}
\end{icmlauthorlist}

\icmlaffiliation{ntua}{School of Electrical and Computer Engineering, National Technical University of Athens, Athens, Attica, Greece}
\icmlaffiliation{bs}{Behavioral Signal Technologies, Los Angeles, CA, USA}

\icmlcorrespondingauthor{Georgios Chochlakis}{georgioschochlakis@gmail.com}

\icmlkeywords{Zero Shot Learning, Few Shot Learning}

\vskip 0.3in
]



\printAffiliationsAndNotice{}  % leave blank if no need to mention equal contribution

\begin{abstract}
    Contemporary state-of-the-art approaches to Zero-Shot Learning (ZSL) train generative
    nets to synthesize examples conditioned on the provided metadata.
    Thereafter, classifiers are trained on these synthetic data in a supervised manner.
    In this work, we introduce \textit{Z2FSL}, an end-to-end generative ZSL framework that
    uses such an approach as a backbone and feeds its synthesized output to
    a Few-Shot Learning (FSL) algorithm. The two modules are trained jointly.
    Z2FSL solves the ZSL problem with a FSL algorithm, reducing,
    in effect, ZSL to FSL. A wide class of algorithms can be
    integrated within our framework. Our experimental results show consistent improvement over several
    baselines. The proposed method, evaluated across standard benchmarks,
    shows state-of-the-art or competitive performance in ZSL and Generalized
    ZSL tasks.
    
    % Contemporary state-of-the-art approaches to Zero-Shot Learning (ZSL) train generative
    % nets to synthesize examples conditioned on the provided metadata.
    % Thereafter, classifiers can be trained on these synthetic data in a supervised manner.
    % In this work, we introduce \textit{Z2FSL}, an end-to-end generative ZSL framework, which
    % uses the generative ZSL pipeline as a backbone and feeds its synthesized output to
    % a Few-Shot Learning (FSL) algorithm. Therefore, the two modules can be trained jointly,
    % \ie{} in an end-to-end manner. Z2FSL solves the ZSL problem with a FSL algorithm, reducing,
    % in effect, ZSL to FSL. A wide class of generative ZSL approaches and FSL algorithms can be
    % integrated within our framework. Our experimental results show consistent improvement over several
    % baseline generative ZSL algorithms. The proposed method, evaluated across standard benchmarks,
    % shows state-of-the-art or competitive performance in ZSL and Generalized
    % ZSL tasks.
   
\end{abstract}

%%%%%%%%%%%%%%%%%%%%%%%%%%%%%%%%%%%%%%%%%%%%%%%%%%%%%%%%%%%%%%%%%%%%%%%%%%%%%%%%%%%%%%%%
%%%%%%%%%%%%%%%%%%%%%%%%%%%%%%%%%%%%%%%%%%%%%%%%%%%%%%%%%%%%%%%%%%%%%%%%%%%%%%%%%%%%%%%%

\section{Introduction}

Deep Learning has seen great success in various settings and disciplines, like
Computer Vision \cite{krizhevsky2012imagenet, zhou2017unsupervised, chen2019learning},
Speech and Language Processing \cite{devlin2018bert, brown2020language}, Computer 
Graphics \cite{starke2019neural, park2019learning} and Medical Science
\cite{ronneberger2015u, rajpurkar2017chexnet}. Despite the variety of applications,
there is a common denominator: resources. The overwhelming majority of applications not
only benefit from but necessitate voluminous data along with the associated hardware 
resources to achieve their reported results. This is evident both from their 
unprecedented, ``superhuman'' performance in many narrow tasks compared to traditional 
algorithms \cite{he2015delving, silver2017mastering} and their surprising deficiencies 
in low-data regimes. Consequently, resource requirements have skyrocketed.
For instance, BERT's~\cite{devlin2018bert} training requirements reach 256 TPU days.

A class of problems that deal with small and medium-sized data sets and distributions shifts are \textbf{Zero-Shot Learning} (\textbf{ZSL}) and \textbf{Few-Shot Learning} (\textbf{FSL}). These
can prove interesting testing grounds for efficient learning. In these problems, and particularly in ZSL
for Computer Vision tasks, Deep Learning has failed to have an immediate impact.
It has only been applied in indirect ways, most notably by replacing image features
extracted with traditional Computer Vision algorithms, like Bag of Visual Words
\cite{weston2010large}, with features extracted by deep nets such as
ResNet~\cite{he2016deep}. Further integration of Deep Learning techniques is
important to advance these settings.

A step towards that direction has been made with the inclusion
of generative models. For FSL, various forms of autoencoders have
been leveraged to provide additional, synthetic examples given the
actual support set \cite{antoniou2017data, wang2018low, xian2019f}.
In ZSL, generative networks conditioned on some form of class
descriptions are trained so as to generate synthetic samples of the test classes.
In this way, the classification task is transformed into a standard supervised
classification task. As a result, classifiers can be trained in a supervised manner
\cite{xian2018feature, zhu2018generative}. More elaborate training techniques
have been proposed \cite{li2019rethinking, xian2019f, li2019leveraging, keshari2020generalized, narayan2020latent}, yet without much progress on extending the basic approach.

% In this work, we build on these generative approaches by including the final
% supervised classifier in the training setting of the generative network,
% rendering the overall process end-to-end. The two modules can be trained
% jointly. The classifier's loss is combined with the prior loss of the
% generative network to form our proposed approach. For the same classifier to
% be viable for both training and testing in a ZSL setting, it
% has to be flexible \wrt{} its output label space and it has to be independent
% of the generative network's synthetic examples. A class of algorithms that
% meets these requirements are Few-shot learners. Therefore, we leverage such
% a classifier by viewing the generator's synthetic examples from a different
% perspective, as a support set rather than a ``standard'' data set, thus
% conceptually reducing ZSL to FSL. 

In this work, we combine these two low-data regimes, namely ZSL and FSL. Specifically, we use the generative ZSL pipeline as a backbone and feed its synthesized output to a FSL classifier. The two modules can be trained jointly, rendering the overall process end-to-end. Formally, the FSL classifier's loss is combined with the prior loss of the generative ZSL framework to form our proposed framework, Z2FSL. Z2FSL conceptually reduces ZSL to FSL by structuring the generator's output as a support set for the FSL algorithm. Using the same FSL classifier during both training and testing is possible because of the flexibility of its output label space. This property holds because the FSL classifier can classify input patterns based on the examples and the classes present in its support set.

Our motivation and rationale for making the process end-to-end is
threefold. First, the generative net gains access to the classification loss of the final
classifier. This is beneficial because sample generation becomes
more discriminative in a manner that explicitly helps
the FSL classifier, since the latter's loss drives the generation. Secondly, thanks to the aforementioned FSL property, the FSL classifier's training is not reliant on the generated samples of the generator,
so the former can be pre-trained on real examples. Additionally, this pre-training is not restricted to the corresponding training set of each task. For example, we can do so on ImageNet~\cite{deng2009imagenet}. Lastly, Few-shot learners perform favorably compared to other alternatives in low-shot classification tasks
\cite{vinyals2016matching, snell2017prototypical, wang2018low}.

Our contributions to the study of ZSL are:
\begin{enumerate}
    \item The coupling of two standard research benchmarks, ZSL and FSL, by
        our novel framework, Z2FSL, which makes generative ZSL approaches
        end-to-end by using a FSL classifier.
    \item Formulating our framework in a manner that allows for a wide class of
        ZSL and FSL algorithms to be seamlessly integrated.
    \item Achieving state-of-the-art or competitive performance on ZSL and Generalized ZSL benchmarks and
        analyzing the contributions of each component of our framework.
\end{enumerate}

We have open-sourced our code\footnote{\url{https://github.com/gchochla/z2fsl}}.

%%%%%%%%%%%%%%%%%%%%%%%%%%%%%%%%%%%%%%%%%%%%%%%%%%%%%%%%%%%%%%%%%%%%%%%%%%%%%%%%%%%%%%%%
%%%%%%%%%%%%%%%%%%%%%%%%%%%%%%%%%%%%%%%%%%%%%%%%%%%%%%%%%%%%%%%%%%%%%%%%%%%%%%%%%%%%%%%%

\section{Related Work}

Earlier works address ZSL by splitting inference into two stages,
inferring the attributes -- the auxiliary description -- of an image and then
assigning the image to the closest given attribute vector. Examples are
\mbox{DAP}~\cite{lampert2013attribute} and the technique presented by
\citet{al2016recovering}. Alternatively, \mbox{IAP}~\cite{lampert2013attribute}
predicts the class posteriors and these are used to calculate the attribute posteriors
of any image. \mbox{Word2Vec}~\cite{mikolov2013efficient} descriptions have also been
used instead of attributes, an example being \mbox{CONSE}~\cite{norouzi2013zero}.

More recent research concentrates on learning a linear mapping from the image-feature 
space to a semantic space. ALE~\cite{akata2015label} learns a compatibility function 
between attributes and image features that is a bilinear form. 
\mbox{SJE}~\cite{akata2015evaluation}, \mbox{DEVISE}~\cite{frome2013devise}, 
\mbox{ESZSL}~\cite{romera2015embarrassingly} and \citet{qiao2016less} learn a bilinear
form as a compatibility function as well. SAE~\cite{kodirov2017semantic} tackles ZSL with
a linear autoencoder. Extending linear mappings,
\citet{xian2016latent} introduced \mbox{LATEM}, which is a piecewise-linear 
compatibility function.

Other recent approaches can be categorized as prototypical, because, at least 
conceptually, a prototype per class is computed.
\mbox{SYNC}~\cite{changpinyo2016synthesized} align graphs in semantic and image-feature
space, calculating prototypes in the process. \mbox{CVCZSL}~\cite{li2019rethinking}
learns a neural net that maps directly from attributes to image features, to class
prototypes that is.

Currently, generative ZSL
stands as the state of the art. The inaugural works of
\citet{xian2018feature, zhu2018generative} laid out the foundation, the basic generative
approach, which can be broken down into three stages: first, a generative network is 
trained to generate instances of seen classes conditioned on the provided class 
descriptions. A differentiable classifier (\eg{} pre-trained linear classifier,
\mbox{AC-GAN}~\cite{odena2017conditional}) can also be used to drive
discriminative generation. In the second stage, given the description of every test 
class, the generative net is used to create a synthetic data set, transforming the 
problem into a supervised one. Then, a supervised classifier (\mbox{SVM}, linear, 
\etc{}) is trained on this data set. In the last stage, the classifier is tasked with
classifying the actual test samples.

This basic approach has been somewhat enriched to improve performance.
\mbox{CIZSL}~\cite{elhoseiny2019creativity} uses creative generation during
training. \mbox{LisGAN}~\cite{li2019leveraging} borrows from prototypical approaches
and utilizes class representatives to anchor generation. 
\mbox{f-VAEGAN}~\cite{xian2019f} shares weights between the decoder of a VAE
and the generator of a WGAN to leverage the better aspects of both.
\mbox{GDAN}~\cite{huang2019generative} uses cycle consistency. 
\mbox{ZSML}~\cite{verma2020meta} introduces meta-learning techniques.
\mbox{OCD}~\cite{keshari2020generalized} uses an over-complete distribution
to generate hard examples and render the synthetic data set more informative.
\mbox{TF-VAEGAN}~\cite{narayan2020latent} uses feedback to augment the f-VAEGAN.

In all cases, the proposed algorithms deviate from the basic approach mainly in 
the training regime of the generative net. In this paper, we describe
Z2FSL, a generative ZSL framework which, via FSL, allows for improvements
to the whole pipeline. In the next section, we describe in detail how this
is achieved.

%%%%%%%%%%%%%%%%%%%%%%%%%%%%%%%%%%%%%%%%%%%%%%%%%%%%%%%%%%%%%%%%%%%%%%%%%%%%%%%%%%%%%%%%
%%%%%%%%%%%%%%%%%%%%%%%%%%%%%%%%%%%%%%%%%%%%%%%%%%%%%%%%%%%%%%%%%%%%%%%%%%%%%%%%%%%%%%%%

\section{Preliminaries}

In this section, we give all the necessary 
definitions, introduce notation and briefly discuss the necessary background.

%%%%%%%%%%%%%%%%%%%%%%%%%%%%%%%%%%%%%%%%%%%%%%%%%%%%%%%%%%%%%%%%%%%%%%%%%%%%%%%%%%%%%%%%
%%%%%%%%%%%%%%%%%%%%%%%%%%%%%%%%%%%%%%%%%%%%%%%%%%%%%%%%%%%%%%%%%%%%%%%%%%%%%%%%%%%%%%%%

\subsection{Problem Definition} \label{sec:prob-def}

Let $\mathcal{X}_{tr}$ be the training samples and $\mathcal{X}_{ts}$ be the test 
samples. Let $\mathcal{Y}_{tr}$ be the corresponding set of training labels and
$\mathcal{Y}_{ts}$ be the corresponding set of test labels.

In the \textbf{ZSL} setting, we have
${\mathcal{Y}_{tr} \cap \mathcal{Y}_{ts} = \emptyset}$, \ie{} there are no common 
classes between training and testing, leading to the terms \textit{seen} and 
\textit{unseen} to describe the classes of the train and test setting respectively.
In the \textbf{Generalized ZSL (GZSL)} setting, that restriction
becomes ${\mathcal{Y}_{tr} \subset \mathcal{Y}_{ts}}$, meaning that there are samples 
from both unseen classes and every seen class during testing. Both in ZSL and 
GZSL, an auxiliary description of each class is provided to counterbalance the absence 
of training examples for unseen test classes. Such a description can be the Word2Vec
representation of the class, an attribute vector or the Wikipedia article for the class.

In a \textbf{FSL} setting, it is sufficient to impose the 
restriction ${\mathcal{Y}_{tr} \subset \mathcal{Y}_{ts}}$, same as GZSL. However, no 
descriptions are provided. Rather, during testing, we are provided with a set of 
examples per test class, named \textit{support set}. A support set essentially consists
of labeled examples from all $n_W$ test classes, with $n_S$ examples each, where $n_W$
and $n_S$ are arbitrary natural numbers referred to as \textit{way} and \textit{shot} 
respectively. Also, let $S_k$ denote the set of examples of class $k$ in a support set.
Given a support set, we are tasked with classifying a set of unlabeled samples of
the same $n_W$ classes called \textit{query set}. For convenience, we consider
query sets to contain $n_Q$  samples per class, which is also an arbitrary natural
number. Randomly sampling a support set and a corresponding query set from a
(usually significantly) larger test data set constitutes an \textit{episode}.
The purpose of an episode is to test the ability of a model to generalize to
possibly unseen classes given only a small number of examples of each class, seen
or unseen, and its classification accuracy on the episode is naturally used as
the metric of success. It is customary to report the average accuracy over
many episodes to capture a more robust metric for a particular data set, where the
way and the shot of the episodes remain constant. That is to say, a specific FSL
setting is characterized by its way and shot and described as $n_W$-\textit{way}
$n_S$-\textit{shot}, \eg{} 25-way 4-shot refers to the regime where the support
sets contain ${n_W = 25}$ classes, with ${n_S = 4}$ samples each.

%%%%%%%%%%%%%%%%%%%%%%%%%%%%%%%%%%%%%%%%%%%%%%%%%%%%%%%%%%%%%%%%%%%%%%%%%%%%%%%%%%%%%%%%
%%%%%%%%%%%%%%%%%%%%%%%%%%%%%%%%%%%%%%%%%%%%%%%%%%%%%%%%%%%%%%%%%%%%%%%%%%%%%%%%%%%%%%%%

\subsection{Background}

\textbf{Wasserstein Generative Adversarial Net}: Wasserstein Generative Neural Nets
(WGAN)~\cite{arjovsky2017wasserstein}, an evolution of Generative Adversarial Nets
\cite{goodfellow2014generative}, are a framework to train a generator $G$, formulating
$p(x|z)$ of real samples $x$ with noise inputs $z$, in a minimax fashion with another network, a
discriminator $D$. \citet{gulrajani2017improved} added a regularization term to
improve performance. We present the formulation of $p(x|a,z)$, which includes a 
conditioning variable $a$, namely
\begin{equation}
    \begin{split}
        \mathcal{L}_{WGAN}&(G, D; p_R, p_Z) = \\
        & \Exp_{(x, a)\sim p_R, z\sim p_Z} \left[ D(x, a) - D(G(a, z), a) \right] \\
        & - \lambda \Exp_{(\hat{x}, a)\sim p_{\hat{X}}} \left[ (\| \nabla_{\hat{x}} D(\hat{x}, a) \|_2 - 1)^2 \right],
    \end{split}
\end{equation}
where $p_R$ is the distribution of the real data, $p_Z$ a ``noise'' distribution, 
$p_{\hat{X}}$ the joint distribution of the conditioning variable and the uniform 
distribution on the line between $x$ and $G(a, z)$, $(x, a) \sim p_R$, $z \sim p_Z$
(intuitively $\hat{x} = ux + (1 - u) G(a, z)$, $u\sim U(0, 1)$) and $\lambda$ is a 
hyperparameter. The minimax game is defined as:
\begin{equation}
    \begin{split}
        \min_G\max_D \mathcal{L}_{WGAN}(G, D; p_R, p_Z).
    \end{split}
\end{equation}
We use \textit{generator} and \textit{generative net} interchangeably.

%%%%%%%%%%%%%%%%%%%%%%%%%%%%%%%%%%%%%%%%%%%%%%%%%%%%%%%%%%%%%%%%%%%%%%%%%%%%%%%%%%%%%%%%
%%%%%%%%%%%%%%%%%%%%%%%%%%%%%%%%%%%%%%%%%%%%%%%%%%%%%%%%%%%%%%%%%%%%%%%%%%%%%%%%%%%%%%%%

\textbf{Variational Autoencoder}: The Variational Autoencoder (VAE), introduced by 
\citet{kingma2013auto}, is an autoencoder that maximizes the variational lower bound of
the marginal likelihood. An autoencoder consists of an encoder $E$, which compresses
its input to a ``latent'' variable $z$, and a decoder $D$, which reconstructs the 
original input of $E$ based on $z$. We present the formulation of the VAE with a 
conditioning variable $a$,
\begin{equation}
    \begin{split}
        \mathcal{L}_{VAE}&(E, D; p_R, p_{\theta}) = \\
        & - \mathbb{E}_{(x,a)\sim p_R}[x\cdot\log{D(E(x, a), a)}\\
        & \qquad + (1-x)\log{(1 - D(E(x, a), a))}] \\
        & + \mathbb{E}_{(x,a)\sim p_R}\left[D_{KL}(p_E(z|x, a)\| p_{\theta}(z))\right],
    \end{split}
\end{equation}
where $D_{KL}$ is the Kullback-Leibler (KL) divergence, $p_R$ the distribution of the
real data, $p_E$ the output distribution of $E$, which is a Gaussian Feedforward
Neural Network (FFNN), \ie{} it outputs the mean and the diagonal elements of the 
covariance of a Gaussian distribution, which makes the KL divergence 
analytical, and $p_{\theta}$ is the prior distribution of the ``latent'' variable. For 
practical purposes, the prior is set to $\mathcal{N}(.\ ; 0, I)$, and the 
reparameterization trick \cite{bengio2013estimating} is used to sample from the encoder as
${z = \mu(x, a) + \epsilon \odot \sigma(x, a)}$, ${\epsilon \sim 
\mathcal{N}(\epsilon; 0, I)}$, where $\odot$ is the Hadamard product and $\mu,\
\sigma$ the encoder's outputs.

%%%%%%%%%%%%%%%%%%%%%%%%%%%%%%%%%%%%%%%%%%%%%%%%%%%%%%%%%%%%%%%%%%%%%%%%%%%%%%%%%%%%%%%%
%%%%%%%%%%%%%%%%%%%%%%%%%%%%%%%%%%%%%%%%%%%%%%%%%%%%%%%%%%%%%%%%%%%%%%%%%%%%%%%%%%%%%%%%

\textbf{f-VAEGAN}: f-VAEGAN \cite{xian2019f} is a generative ZSL approach that
deploys both a WGAN and a VAE to train the generator, by sharing its weights between
the VAE's decoder and the WGAN's generator. The overall loss function of the approach is
\begin{equation} \label{eq:vaegan}
    \begin{split}
        \mathcal{L}_{VAEGAN}&(G, E, D; p_R, p_Z, \beta) = \\
        & \mathcal{L}_{VAE}(E, G; p_R, p_Z) \\
        & + \beta \cdot \mathcal{L}_{WGAN}(G, D; p_R, p_Z),
    \end{split}
\end{equation}
where $E$ is the encoder of the VAE, $D$ the discriminator of the WGAN, $G$ the generator 
of the WGAN and the decoder of the VAE and $\beta$ a hyperparameter.

%%%%%%%%%%%%%%%%%%%%%%%%%%%%%%%%%%%%%%%%%%%%%%%%%%%%%%%%%%%%%%%%%%%%%%%%%%%%%%%%%%%%%%%%
%%%%%%%%%%%%%%%%%%%%%%%%%%%%%%%%%%%%%%%%%%%%%%%%%%%%%%%%%%%%%%%%%%%%%%%%%%%%%%%%%%%%%%%%

\textbf{Prototypical Network}: Prototypical Networks (PN) \cite{snell2017prototypical} 
present a simple, differentiable framework for FSL. A neural network, $f_{\phi}$, is
used to map the input samples to a metric space. The support set is mapped and the
embeddings are averaged per class so as to get a prototype $c_k$ for each. Then,
each sample in the query set is mapped to the metric space and classified to the nearest
prototype based on the Euclidean distance $d(\cdot \ , \cdot)$. The formulation is
\begin{equation} \label{eq:pn}
    \begin{split}
        c_k =\ & \frac{1}{|S_k|}\sum_{x_i\in S_k} f_{\phi}(x_i), \\
        p_{\phi}(y=k|x) =\ & \frac{\exp{(-d(f_{\phi}(x), c_k))}}{\sum_{k'}\exp{(-d(f_{\phi}(x), c_{k'}))}}, \\
        \mathcal{L}_{PN}(f_{\phi}; S, Q) =\ & \frac{1}{|Q|}\sum_{(x_i, y_i)\in Q}\log{p_{\phi}(y=y_i|x)}, \\
    \end{split}
\end{equation}
where $p_{\phi}(y|x)$ is the softmax output distribution of $x$ belonging to some
class.  To compute $\mathcal{L}_{PN}$, we have to sample episodes instead of batches. In
this manner, training resembles testing. We refer to that manner of training
as \textit{episodic}.

%%%%%%%%%%%%%%%%%%%%%%%%%%%%%%%%%%%%%%%%%%%%%%%%%%%%%%%%%%%%%%%%%%%%%%%%%%%%%%%%%%%%%%%%
%%%%%%%%%%%%%%%%%%%%%%%%%%%%%%%%%%%%%%%%%%%%%%%%%%%%%%%%%%%%%%%%%%%%%%%%%%%%%%%%%%%%%%%%

\begin{figure*}[ht]
\vskip 0.2in
\begin{center}
\centerline{\includegraphics[width=1.8\columnwidth]{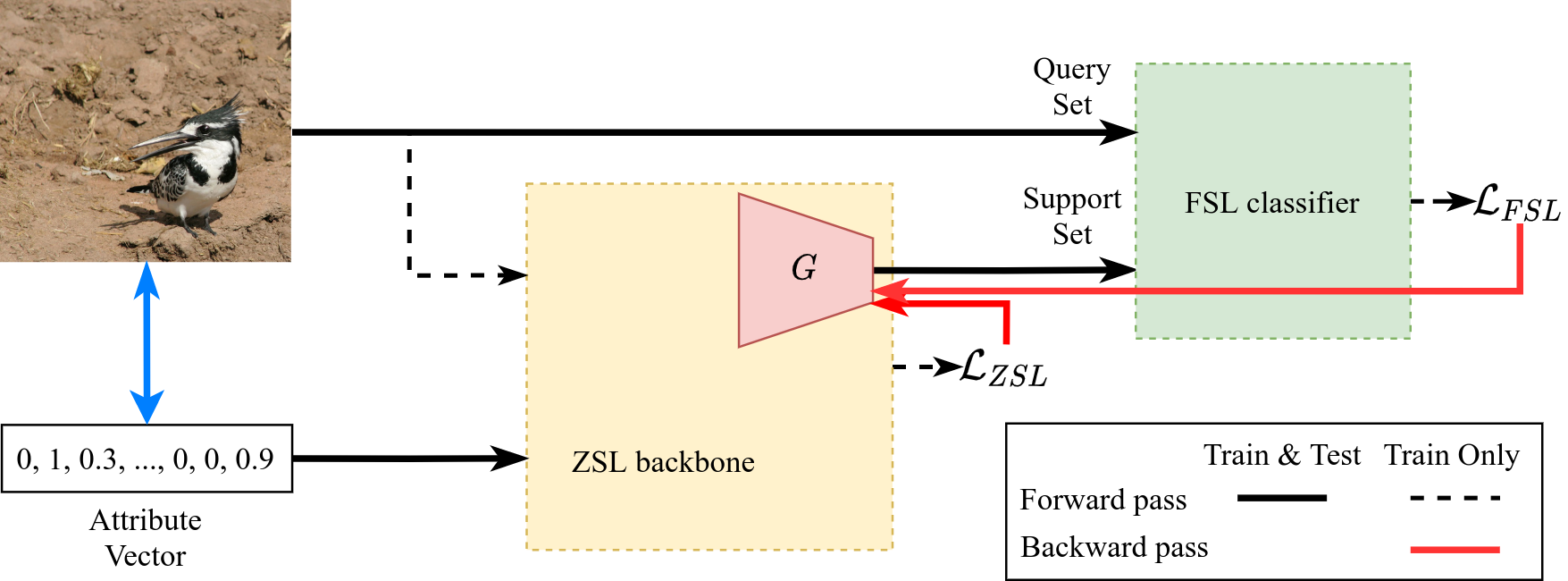}}
\caption{Graphical representation of the Z2FSL framework and pipeline. We use a
generative Zero-shot Learning backbone with a Few-shot Learning classifier. During
training, we train the generator $G$ with a combination of its training within the
backbone and the Few-shot Learning classification task. During testing, the forward
pass is only altered in that real examples are not provided to the backbone and
no backward pass is performed.}
\label{fig:z2fsl}
\end{center}
\vskip -0.2in
\end{figure*}

\section{Method}

% \efthygeo{We now describe the proposed Z2FSL framework in detail. In Z2FSL a generative ZSL pipeline is coupled with a FSL classifier. Specifically, the generator network is used to synthetize the support set for the FSL classifier. We can train the two modules
% jointly, conceptually reducing ZSL to FSL, by generating the support set of the FSL classifier via the ZSL generator. Additionally, both modules can be separately pre-trained on other data and further fine-tuned through our framework. + add refs to Figure 1}

We now describe the proposed Z2FSL framework in detail. In Z2FSL,
a generative ZSL pipeline, used as a backbone, is coupled with a FSL classifier.
We can train the two modules jointly by conceptually reducing ZSL to FSL, which simply means that the backbone generates the classifier's support set. To get an episode, real examples of the classes that are present
in this support set can be used as the query set. During testing, the test samples become the query and the support is again provided by the backbone. The framework is also presented in Figure \ref{fig:z2fsl}, where a graphical representation of how the novel component of our framework, the FSL classifier, affects the pipeline can be seen. The backbone provides the support set of the FSL classifier in all settings. During training, this allows the back-propagation of the FSL loss to the generator. During testing, synthesized examples of the test classes are provided in order to enable the classification of unseen classes.

This is possible because the FSL algorithm can classify its input dynamically, in
the sense that its output distribution is based on the classes present in the support
set. This allows us to
train the classifier on classes other than the unseen classes and simply provide
the necessary synthetic support set during testing. Another advantage is that the FSL algorithm 
can actually be pre-trained before
the joint training with the backbone and/or fine-tuned afterwards, or even trained completely separately of the backbone.

\subsection{Training}

% (\eg{} FSL algorithm, generator, discriminator),
We train all the components for multiple iterations. As an example, the components in this work, other than the generator and the FSL classifier, which are also visible in Figure~\ref{fig:z2fsl}, include the discriminator of a WGAN and the encoder of a VAE. In each iteration, we take steps training the components:
training the FSL classifier, training the generator and training all the
other components as necessary.

First, when training the FSL algorithm, we randomly sample query sets from the training
set and support sets are generated by the backbone, conditioned on the metadata of the classes
that appear in the query set. The FSL loss, denoted $\mathcal{L}_{FSL}$, remains
intact (\eg{} Equation~\ref{eq:pn} for PNs). This is a slightly modified version
of the episodic training we defined in Section~\ref{sec:prob-def}, since the support set is now synthetic. Episodes during
training, $n_W$, $n_S$ and $n_Q$ in particular, can be set arbitrarily, similar to
batches. The process can be seen indirectly in Figure~\ref{fig:z2fsl}. For this step, we would have to simply back-propagate $\mathcal{L}_{FSL}$ only to the FSL classifier.

Second, when training the generator of the ZSL backbone, we use the loss within the generative framework
of the backbone, which we denote as $\mathcal{L}_{ZSL}$ (\eg{} Equation~\ref{eq:vaegan} if the backbone is the f-VAEGAN). By feeding the generated samples to the FSL algorithm, we can back-propagate
$\mathcal{L}_{FSL}$ to the generator. This also requires sampling a query set based on the
classes the generator provides. The overall objective of the
ZSL generator is then described as:
\begin{equation} \label{eq:z2fsl}
    \mathcal{L}_{Z2FSL} = \mathcal{L}_{ZSL} + \gamma \mathcal{L}_{FSL},
\end{equation}
where $\gamma$ is a hyperparameter. This is the step depicted in Figure~\ref{fig:z2fsl} if we take into account all arrows.

The rest of the components can be trained before or after the two aforementioned steps.
In this work, for example, we update the discriminator multiple times before
each generator update, as suggested by \citet{goodfellow2014generative, arjovsky2017wasserstein}.

However, components of the backbone can even be trained along with the generator. A component that
does not affect the generation of the support set in the forward pass can be trained along with the generator, but it remains unaffected by $\mathcal{L}_{FSL}$. Such a component is the encoder of a VAE,
which we also use in our work, or a regressor for cycle consistency \cite{huang2019generative}. A component that affects generation in the forward pass, such as a feedback module \cite{narayan2020latent},
can be trained along with the generator and updated based on the objective in Equation~\ref{eq:z2fsl} rather than $\mathcal{L}_{ZSL}$ alone.

In Figure~\ref{fig:z2fsl}, we can more generally see that both real examples and corresponding descriptions are provided to the backbone, while only real and generated examples to the classifier. Real samples are useful to the backbone only during training, \eg{} to compute the VAE reconstruction loss.

% Other components can be trained before or after the aforementioned steps, or
% trained along with the generator. For example, a discriminator can be trained
% for multiple steps before the generator step
% \cite{goodfellow2014generative, arjovsky2017wasserstein}. A component that
% affects generation (in the forward pass), such as a feedback module
% \cite{narayan2020latent}, can be trained along with the generator and updated
% based on the objective in Equation \ref{eq:z2fsl} rather than $\mathcal{L}_{ZSL}$
% alone. A component that does not affect generation (in the forward pass), like
% a regressor for cycle consistency \cite{huang2019generative}, can obviously be
% trained along with the generator, but it remains unaffected by $\mathcal{L}_{FSL}$.

% The pipeline can also be seen in Figure \ref{fig:z2fsl}. We see that both samples
% and corresponding descriptions are provided to the backbone, while only real and
% generated samples to the classifier. Real samples are useful to the backbone only
% during training, \eg{} to calculate the VAE reconstruction loss.

\subsection{Evaluation}

% \efthygeo{During evaluation stage we stop providing real samples to the backbone. explain what happens. e/g the zsl backbone generates instances solely based on the linguistic description/ attribute vector...
% kai meta auta ta akatalavistika me $n_w$}

For the evaluation, the pipeline is modified in two ways. First, as shown
in Figure~\ref{fig:z2fsl}, we stop providing real samples to the backbone. Only the descriptions are necessary so as to generate the test support set. Second, episodes are restricted by the test setting. The number of classes in each set,
$n_W$, is fixed and equal to the number of test classes. $n_Q$ is equal, for each test class,
to the number of test samples available. $n_S$ remains a hyperparameter, as we can
choose the number of samples to generate per class. In this manner, the query set
contains all test samples and the FSL classification accuracy is exactly the final ZSL accuracy.

\subsection{Assumptions}

We make three assumptions about the backbone and the classifier altogether. First, the
backbone is a generative ZSL approach. Second, the classifier is trained through a 
differentiable process. Third, the classifier can classify its input dynamically, by 
matching support and query classes
\cite{vinyals2016matching, snell2017prototypical, wang2018low}. This fact allows the 
usage of the classifier during testing without any training on unseen classes as long
as corresponding supporting examples are provided at that time. This means that the 
classifier is not reliant on the generator and, consequently, the modules can be trained separately or jointly.

As a result, our framework is suitable for a wide class of ZSL and FSL algorithms. We 
formulate our framework as agnostic to the generative ZSL and the FSL algorithm, and 
refer to a specific implementation of it by the following macro: Z2FSL(\texttt{z}, \texttt{f}), where \texttt{z} is the generative ZSL backbone and \texttt{f} the FSL classifier.

%%%%%%%%%%%%%%%%%%%%%%%%%%%%%%%%%%%%%%%%%%%%%%%%%%%%%%%%%%%%%%%%%%%%%%%%%%%%%%%%%%%%%%%%
%%%%%%%%%%%%%%%%%%%%%%%%%%%%%%%%%%%%%%%%%%%%%%%%%%%%%%%%%%%%%%%%%%%%%%%%%%%%%%%%%%%%%%%%

\section{Experiments}

We first present the datasets, followed by implementation details and finally present
our experimental results along with comparison to state ot the art.

\subsection{Data sets} \label{sec:datasets}

We use the Caltech UCSD Bird 200 (CUB, \citealt{wah2011caltech}), which consists
of 11788 images of birds belonging to 200 species. One
312-dimensional attribute vector per class is provided as well. We also use Animals with Attributes 2 (AwA2,
\citealt{xian2018zero}). It contains 37322 images from 50 categories and 85-dimensional
attributes. Our last data set is the SUN Scene Classification (SUN, \citealt{patterson2012sun})
data set, with 14340 images of 717 categories and 102-dimensional attributes.

We use the provided attributes as our auxiliary descriptions, and in particular the 
continuous attributes after normalizing them \wrt{} their L2 norm. We use 10 crops per
image (original image, top-right, top-left, bottom-right, bottom-left and their
horizontally flipped counterparts) as augmentation. We use the original image for
testing. Instead of images, we use features extracted by the
\mbox{ResNet-101}~\cite{he2016deep} trained on ImageNet~\cite{deng2009imagenet}.
We choose the 2048-dimensional output of its adaptive average pooling layer.
Additionally, we perform min-max normalization of the features to $[0, 1]$ and use
the train-test and seen-unseen splits proposed by \citet{xian2018zero}.

\begin{figure}[ht]
\vskip 0.2in
\begin{center}
\centerline{\includegraphics[width=\columnwidth]{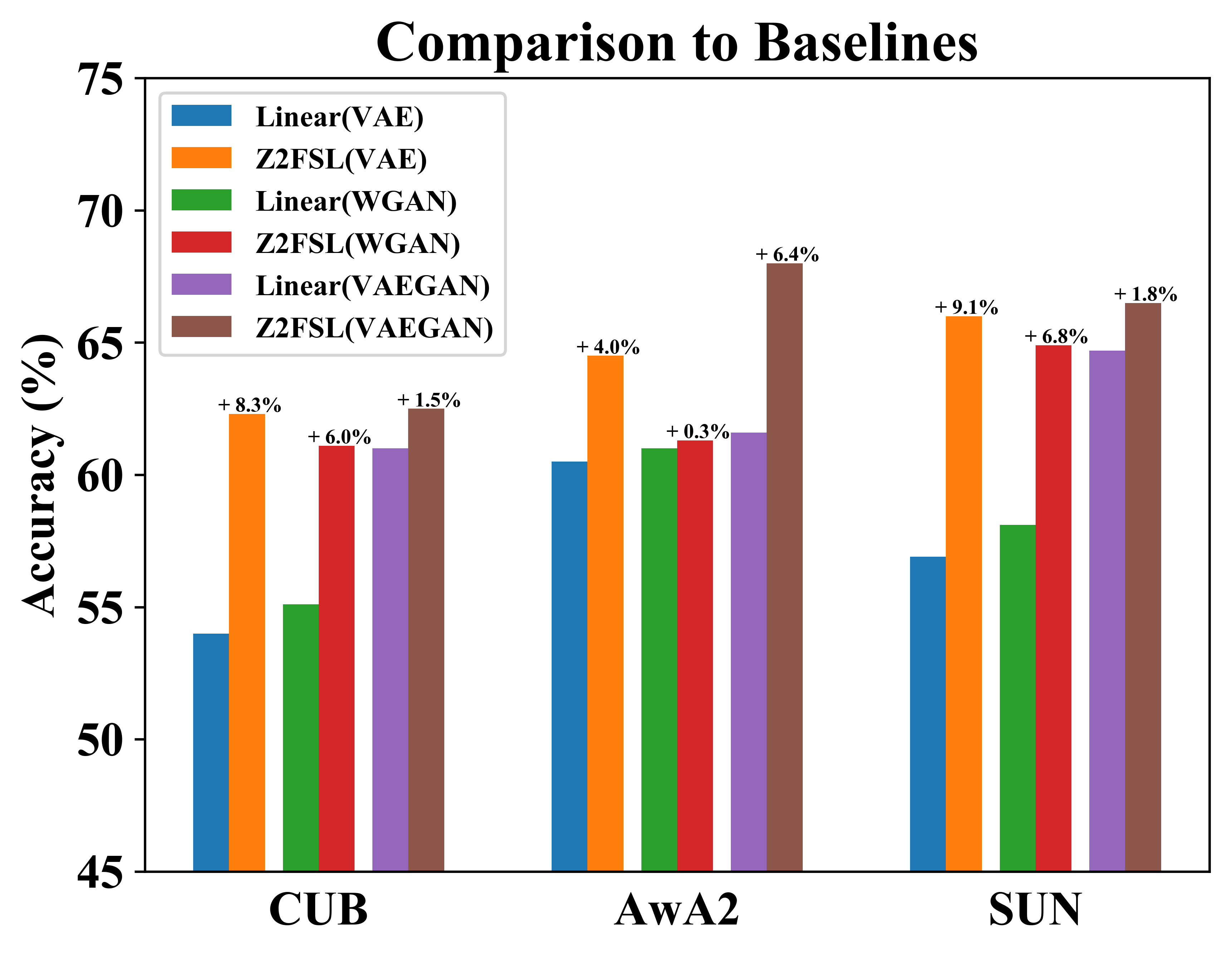}}
\caption{\textbf{Comparison of Z2FSL to baseline generative ZSL backbones}. Linear(\texttt{z})
describes the setting where the generative ZSL approach \texttt{z} is used with a linear classifier trained on the generated data. We skip the second argument of the Z2FSL macro for convenience, since it is the PN in all cases. Linear(VAEGAN) results for CUB and SUN reported from \citet{xian2019f}. The metric presented is the average per class top-1 accuracy. Performance gains denote absolute improvement compared to the corresponding baseline.}
\label{fig:baseline}
\end{center}
\vskip -0.2in
\end{figure}

\subsection{Implementation Details} \label{sec:details}

Since we present the results of \mbox{Z2FSL(f-VAEGAN, PN)} in comparison to the
state of the art, we present details for that specific architecture in this section.

We pre-train the PN in an episodic manner as suggested by \citet{snell2017prototypical},
where support and query sets are sampled from the real training data of the
corresponding data set. The PN is implemented as a FFNN with $n_h$ hidden layers
with square weight  matrices and ReLU activations. The FSL classifier's learning
rate is kept the same in pre-training and the joint training with the generator.

The generator and the encoder are FFNNs with 2 hidden layers each, 4096 followed by
8192 units for the generator and the reverse for the encoder, Leaky ReLU hidden 
activations (0.2 slope), linear output for the encoder and sigmoid for the generator. 
The noise dimension is chosen to be equal to the dimension of the attributes. The 
discriminator is a FFNN with one hidden layer of 4096 neurons and Leaky ReLU hidden 
activations (0.2 slope). We set the coefficient of the regularization term of WGAN 
$\lambda = 10$ and the training updates of the discriminator per generator update equal to 
5.

When we sample support sets from the generative
net, we set $n_S = 5$ during training. During testing, we set it to $n_S = 1800$ for unseen classes. For seen classes in the GZSL test setting, we experiment with both seen and unseen support and select the better alternative for each benchmark. We also consider the shot of the test support set for seen classes a different hyperparameter $m_S$.

The optimizer in all cases is Adam~\cite{kingma2014adam} with $\beta_1 = 0.5$
and $\beta_2 = 0.999$. We apply gradient clipping, restricting the gradient for each
parameter within $[-5, 5]$. Our implementation is in PyTorch~\cite{paszke2019pytorch}.

The tunable hyperparameters that we either search for or vary by data set are presented
in the Supplementary Material.

\begin{table}[t]
\caption{Comparison of our approach, \mbox{Z2FSL(f-VAEGAN, PN)}, to previous work. The metric presented is the average per class top-1 accuracy.}
\label{tbl:zsl}
\vskip 0.15in
\begin{center}
\begin{small}
\begin{sc}
\begin{tabular}{lccc}
\toprule
 & \multicolumn{3}{c}{Zero-shot Learning}  \\
\cmidrule(lr){2-4}
Approach & \textbf{CUB} & \textbf{AwA2} & \textbf{SUN} \\
\midrule
CVCZSL \cite{li2019rethinking} & 54.4 & 71.1 & 62.6 \\
f-CLSWGAN {\scriptsize\cite{xian2018feature}} & 57.3 & - & 60.8 \\
LisGAN \cite{li2019leveraging} & 58.8 & - & 61.7 \\
f-VAEGAN \cite{xian2019f} & 61.0 & - & 64.7 \\
OCD \cite{keshari2020generalized} & 60.3 & 71.3 & 63.5 \\
TF-VAEGAN {\scriptsize\cite{narayan2020latent}} & \textbf{64.9} & \textbf{72.2} & 66.0 \\
Z2FSL(f-VAEGAN, PN) & 62.5 & 68.0 & \textbf{66.5} \\
\bottomrule
\end{tabular}
\end{sc}
\end{small}
\end{center}
\vskip -0.1in
\end{table}

\subsection{Backbone Baselines} \label{sec:baselines}

In this section, we present the performance of our framework using various generative ZSL
approaches as backbones and compare that with the plain generative ZSL approaches, \ie{} without the FSL algorithm, as baselines. The comparison can be seen in Figure~\ref{fig:baseline}. The baselines we have chosen are a VAE, a WGAN and a f-VAEGAN. Z2FSL improves the performance of all baselines consistently across all benchmarks. It is interesting to see that, in most cases, the simple backbones, the VAE and the WGAN, enhanced by our framework, exceed the performance of the more elaborate and superior -- on its own -- plain f-VAEGAN. 

\subsection{State of the art}

In this section, we present our results in comparison with the current state-of-the-art approaches that use the same test setting as our approach (feature extractor, dimensionality of features, splits, \etc{} described in Sections~\ref{sec:datasets} and \ref{sec:details}).

\begin{table*}[t]
\caption{Comparison of our approach, \mbox{Z2FSL(f-VAEGAN, PN)}, to previous work. The metrics presented are: $\mathbf{u}$ is the average per class top-1 accuracy of unseen classes, $\mathbf{s}$ is the average per class top-1 accuracy of seen class and $\mathbf{H}$ their harmonic mean. $\mathbf{H}$ is considered the main metric of this setting.}
\label{tbl:gzsl}
\vskip 0.15in
\begin{center}
\begin{small}
\begin{sc}
\begin{tabular}{l*{9}{c}}
\toprule
 & \multicolumn{9}{c}{Generalized Zero-shot Learning}  \\
\cmidrule(lr){2-10}
& \multicolumn{3}{c}{\textbf{CUB}} & \multicolumn{3}{c}{\textbf{AwA2}} & \multicolumn{3}{c}{\textbf{SUN}} \\
\cmidrule(lr){2-4} \cmidrule(lr){5-7} \cmidrule(lr){8-10} 
Approach & \textbf{u} & \textbf{s} & \textbf{H} & \textbf{u} & \textbf{s} & \textbf{H} & \textbf{u} & \textbf{s} & \textbf{H} \\
\midrule
CVCZSL \cite{li2019rethinking} & 47.4 & 47.6 & 47.5 & 56.4 & \textbf{81.4} & 66.7 & 36.3 & 42.8 & 39.3 \\
f-CLSWGAN {\cite{xian2018feature}} & 43.7 & 57.7 & 49.7 & - & - & - & 42.6 & 36.6 & 39.4 \\
LisGAN \cite{li2019leveraging} & 46.5 & 57.9 & 51.6 & - & - & - & 42.9 & 37.8 & 40.2 \\
f-VAEGAN \cite{xian2019f} & 48.4 & 60.1 & 53.6 & - & - & - & 45.1 & 38.0 & 41.3 \\
OCD \cite{keshari2020generalized} & 44.8 & 59.9 & 51.3 & 59.5 & 73.4 & 65.7 & 44.8 & \textbf{42.9} & \textbf{43.8} \\
TF-VAEGAN {\cite{narayan2020latent}} & \textbf{52.8} & \textbf{64.7} & \textbf{58.1} & \textbf{59.8} & 75.1 & 66.6 & \textbf{45.6} & 40.7 & 43.0 \\
Z2FSL(f-VAEGAN, PN) & 47.2 & 61.2 & 53.3 & 57.4 & 80.0 & \textbf{66.8} & 44.0 & 32.9 & 37.6 \\
\bottomrule
\end{tabular}
\end{sc}
\end{small}
\end{center}
\vskip -0.1in
\end{table*}

\subsubsection{Zero-shot Learning}

For ZSL, our experimental evaluation in Section \ref{sec:baselines} shows that \mbox{Z2FSL(f-VAEGAN, PN)} outperforms the \mbox{f-VAEGAN}. This performance compares favorably to the rest of the state-of-the-art approaches as well, as can be seen in Table~\ref{tbl:zsl}. In particular, even though the \mbox{TF-VAEGAN} itself builds on top of the \mbox{f-VAEGAN} and improves performance, our approach outperforms that on SUN, where it achieves state-of-the-art performance, improving the previous one by an absolute margin of $0.5\%$.

\subsubsection{Generalized Zero-shot Learning}

For GZSL, results can be seen in Table~\ref{tbl:gzsl}. Performance in AwA2 is marginally better than the previous state of the art, \mbox{CVCZSL} and \mbox{TF-VAEGAN}. In contrast, in CUB and SUN it is harder to balance seen and unseen accuracies, as decreasing the $s$ metric, which is partly controlled by $m_S$, is required to achieve the best $H$ possible. This leads to inferior performance compared to our specific backbone in SUN and marginally worse accuracy in CUB. The state-of-the-art results in AwA2 can be partly explained by the small number of classes, which are 50 in total, only 10 of which are unseen.

\begin{figure}[ht]
\vskip 0.2in
\begin{center}
\centerline{\includegraphics[width=\columnwidth]{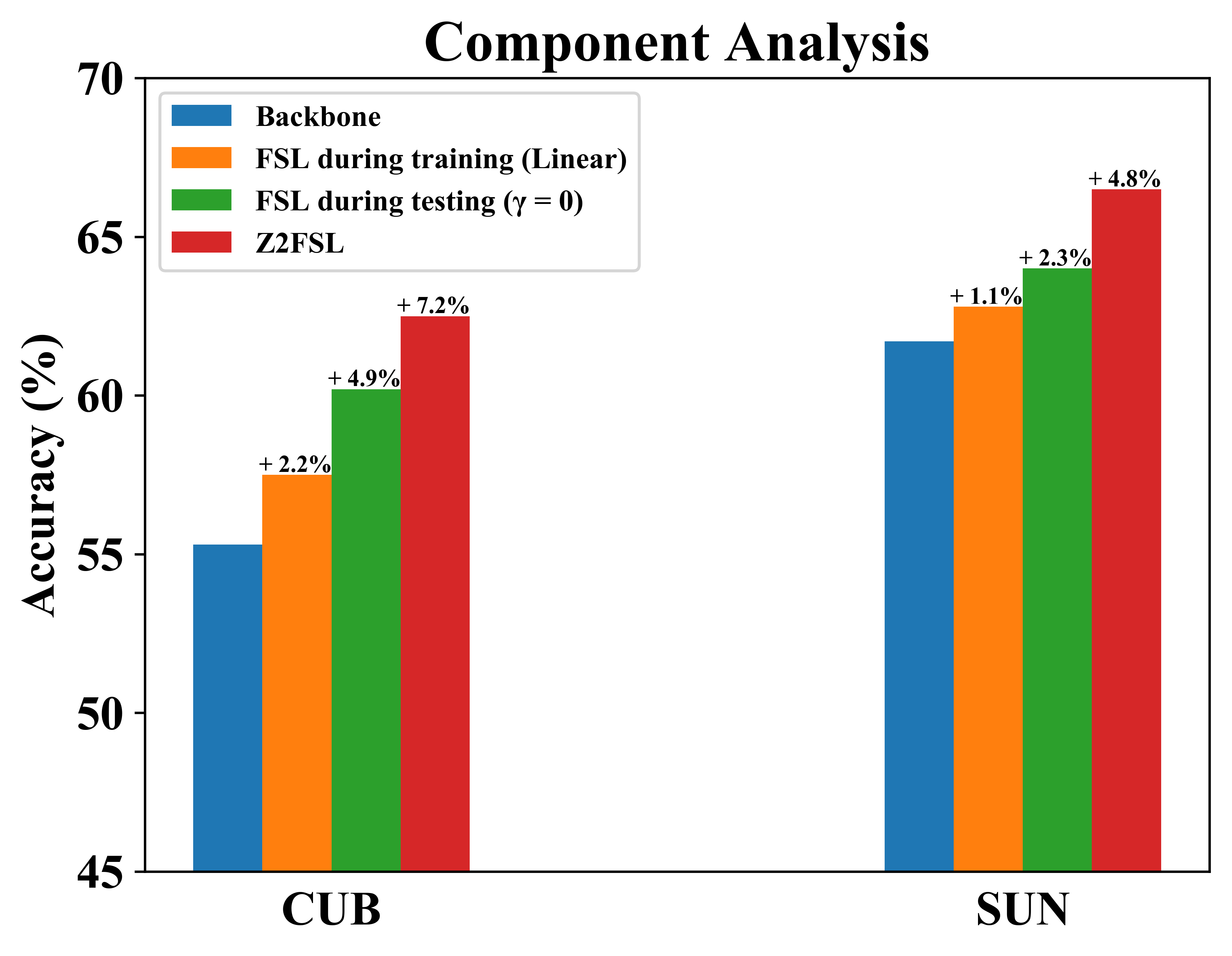}}
\caption{\textbf{FSL classifier's effects on performance}. We compare the performance of the \mbox{f-VAEGAN}, the backbone generative ZSL approach, to the performance of \mbox{Z2FSL(f-VAEGAN, PN)} when we discard the FSL classifier during the evaluation and use a linear classifier instead, the performance of \mbox{Z2FSL(f-VAEGAN, PN)} when $\gamma = 0$ (Equation~\ref{eq:z2fsl}), \ie{} discarding the FSL classifier during training only, and the complete \mbox{Z2FSL(f-VAEGAN, PN)} (with the FSL classifier in both settings). The metric presented is the average per class top-1 accuracy. Performance gains denote absolute improvement compared to the backbone.}
\label{fig:component}
\end{center}
\vskip -0.2in
\end{figure}

\begin{table*}[t]
\caption{Comparison of our approach, \mbox{Z2FSL(f-VAEGAN, PN)} with real and synthetic support for seen classes. The metrics presented are: $\mathbf{u}$ is the average per class top-1 accuracy of unseen classes, $\mathbf{s}$ is the average per class top-1 accuracy of seen class and $\mathbf{H}$ their harmonic mean.}
\label{tbl:real-support}
\vskip 0.15in
\begin{center}
\begin{small}
\begin{sc}
\begin{tabular}{l*{9}{c}}
\toprule
 & \multicolumn{9}{c}{Generalized Zero-shot Learning}  \\
\cmidrule(lr){2-10}
& \multicolumn{3}{c}{\textbf{CUB}} & \multicolumn{3}{c}{\textbf{AwA2}} & \multicolumn{3}{c}{\textbf{SUN}} \\
\cmidrule(lr){2-4} \cmidrule(lr){5-7} \cmidrule(lr){8-10} 
Approach & \textbf{u} & \textbf{s} & \textbf{H} & \textbf{u} & \textbf{s} & \textbf{H} & \textbf{u} & \textbf{s} & \textbf{H} \\
\midrule
Z2FSL(f-VAEGAN, PN) (with real support)  & 47.2 & 61.2 & \textbf{53.3} & 49.4 & 77.7 & 60.4 & 48.1 & 27.4 & 34.9 \\
Z2FSL(f-VAEGAN, PN) & 44.4 & 58.0 & 50.3 & 57.4 & 80.0 & \textbf{66.8} & 44.0 & 32.9 & \textbf{37.6} \\
\bottomrule
\end{tabular}
\end{sc}
\end{small}
\end{center}
\vskip -0.1in
\end{table*}

\subsection{Ablation studies}

\subsubsection{Component Analysis}

We perform a more detailed analysis of how the usage of the novel component of our approach, the FSL classifier, affects performance. We experiment with using the FSL classifier solely during the training of the backbone. During testing, we follow the standard practise of training a linear classifier on the synthetic data. We also experiment with using the FSL classifier only during testing, \ie{} simply setting $\gamma = 0$ in Equation~\ref{eq:z2fsl}. Results are presented in Figure~\ref{fig:component}. We can observe that both regimes yield improvement compared to the plain backbone. Additionally, regarding the gains in performance compared to the backbone, the gain of Z2FSL is greater than the sum of the gains of the two ablation studies. This shows that end-to-end training yields a significant improvement and validates that this process renders the generation discriminative in a manner that explicitly helps the classifier.
%Additionally, it is interesting to note that the absolute gains in performance -- compared to the backbone -- when the FSL algorithm is missing from only one setting, when summed, are still less than the overall gain. This validates the importance of end-to-end training which renders the generation discriminative in a manner that explicitly accommodates the classifier. 

\begin{table}[t]
\caption{Comparison of our approach, \mbox{Z2FSL(f-VAEGAN, PN)}, with and without pre-training the FSL classifier. The metric presented is the average per class top-1 accuracy.}
\label{tbl:pre-training}
\vskip 0.15in
\begin{center}
\begin{small}
\begin{sc}
\begin{tabular}{lcc}
\toprule
 & \multicolumn{2}{c}{ZSL}  \\
\cmidrule(lr){2-3}
Approach & \textbf{CUB} & \textbf{SUN} \\
\midrule
Z2FSL {\small (no pre-training)} & 58.0  & 61.3 \\
Z2FSL & \textbf{62.5} & \textbf{66.5} \\
\bottomrule
\end{tabular}
\end{sc}
\end{small}
\end{center}
\vskip -0.1in
\end{table}

\subsubsection{Synthetic vs. Real Support Set} \label{sec:support-abl}

We also examine the performance of Z2FSL when using real supporting examples of seen classes compared to performance with synthetic ones. Results are presented in Table~\ref{tbl:real-support}. In AwA2 and SUN, the synthetic support results in a better harmonic mean. On the other hand, in CUB, real support leads to an increase in performance. Moreover, as we noted in Section~\ref{sec:details}, the shot for seen classes in the test support set is controlled by a different hyperparameter than that of unseen classes, $m_S$ and $n_S$ (for testing) respectively \footnote{The results in Table~\ref{tbl:real-support} are achieved with a $n_S$ that is roughly two orders of magnitude greater than $m_S$}. These facts clearly demonstrate the bias towards seen classes the naive approaches of using all the available training data in the support set or too many synthetic ones could lead to.

\subsubsection{Pre-training}

We also test the effects of pre-training the FSL classifier of our framework. It makes sense, intuitively, to pre-train it in an actual FSL setting before the joint training with the generator, where we choose to train it with a combination of real and synthetic data. Table~\ref{tbl:pre-training} shows a decrease in performance without pre-training, 5.2\% absolute decrease in SUN and 4.5\% in CUB to be exact. This illustrates that training in the FSL setting is essential to avoid overfitting.

\section{Conclusions}

In this paper, we introduce a novel, end-to-end generative ZSL framework, Z2FSL. Z2FSL uses the same supervised classifier during both training and testing. We choose a FSL algorithm to fill that role, since the choice of classifier is restricted by the ZSL setting. In this manner, we also couple the two low-data regimes. We formulate our framework so as to allow a broad class of generative ZSL approaches and FSL classifiers to be integrated. Empirically, we show that our framework improves upon the results of the plain generative approach. Extensive ablation studies reveal that the improvement originates from the fact that the generated samples of the generative net are rendered more discriminative in a way that explicitly helps the classifier. In addition, these studies demonstrate the advantages of being able to use a pre-trained classifier. Our results are state of the art or competitive across all benchmarks.
We also show that using synthetic samples for seen classes as well as decreasing the number of samples of these classes compared to unseen ones can improve performance in GZSL. 

Our future plans include further investigating and mitigating the bias towards seen classes in the GZSL. 
% Second, some short of fine-tuning after the joint training of the FSL algorithm and the generative net could further improve performance, \eg{} fine-tuning the FSL algorithm on unseen classes by generating both the support and the query set.
Another research direction we plan to investigate are techniques to better train the FSL classifier separately of the generator. Initial thoughts include consistent fine-tuning on generated samples of unseen classes after the joint training and extensive pre-training on other data sets.
%Third, it is now possible to train the generative net only to facilitate the classifier and not within a generative framework. More research towards that could prove fruitful.
Since we now have an end-to-end process to train the generative net, it is possible to completely dispose of the generative framework and, therefore, we also plan to investigate this training regime.

\bibliography{references}
\bibliographystyle{icml2021}

\clearpage

\begin{table*}[t]
\caption{Hyperparameter configuration per setting and data set for the Prototypical
	Network's pre-training. From top to bottom, the hyperparameters
	presented are the learning rate of the FSL algorithm $\alpha_h$, the
	number of episodes $N_h$, the number of hidden layers $n_h$, the number
	of classes in an episode $n_W$, the number of support examples per
	class $n_S$, and the number of queries per class $n_Q$.}
\label{tbl:hyperparameters_pretrain}
\vskip 0.15in
\begin{center}
\begin{small}
\begin{sc}
\begin{tabular}{lcccccc}
	    \toprule
	    & \multicolumn{3}{c}{\textbf{ZSL}} &
	        \multicolumn{3}{c}{\textbf{GZSL}} \\
	    \cmidrule(lr){2-4} \cmidrule(l){5-7}
		Hyperparameter & CUB & AwA2 & SUN & CUB & AwA2 & SUN \\
		\midrule
	    $\alpha_h$ & $5\cdot 10^{-5}$ & $10^{-5}$ & $10^{-5}$ &
	        $10^{-3}$ & $10^{-5}$ & $5\cdot 10^{-5}$ \\
	    $N_h$ & 12000 & 15000 & 10000 & 10000 & 12000 & 8000 \\
	    $n_h$ & 0 & 1 & 1 & 0 & 1 & 1 \\
	    $n_W$ & 25 & 10 & 40 & 25 & 10 & 50 \\
	    $n_S$ & 5 & 5 & 5 & 5 & 5 & 5 \\
	    $n_Q$ & 10 & 15 & 5 & 10 & 15 & 2 \\
		\bottomrule
\end{tabular}
\end{sc}
\end{small}
\end{center}
\vskip -0.1in
\end{table*}

\begin{table*}[t]
\caption{Hyperparameter configuration per setting and data set for our main experiments.
	We have the ZSL learning rate $\alpha_f$, the
	coefficient of the WGAN loss $\beta$, the coefficient of the FSL loss $\gamma$,
	the number of classes in a training episode $n_W$, the number of generations
	per class in during training $n_S$, the number of generations per seen class
	during testing $m_S$, the number of ``queries'' per class in a training
	episode $n_Q$, and the number of episodes $N$.}
\label{tbl:hyperparameters}
\vskip 0.15in
\begin{center}
\begin{small}
\begin{sc}
\begin{tabular}{lcccccc}
	    \toprule
	    & \multicolumn{3}{c}{\textbf{ZSL}} & 
	        \multicolumn{3}{c}{\textbf{GZSL}} \\
	    \cmidrule(lr){2-4} \cmidrule(l){5-7}
		Hyperparameter & CUB & AwA2 & SUN & CUB & AwA2 & SUN \\
		\midrule
		$\alpha_f$ & $10^{-4}$ & $10^{-4}$ & $10^{-4}$ &
		    $10^{-4}$ & $10^{-4}$ & $10^{-4}$ \\
		$\beta$ & 100 & 100 & 100 & 100 & 100 & 100 \\
		$\gamma$ & 100 & 100 & 100 & 10 & 10 & 10 \\
		$n_W$ & 25 & 10 & 80 & 25 & 10 & 80 \\
		$n_S$ & 5 & 5 & 5 & 5 & 5 & 5 \\
        $m_S$ & - & - & - & 5 & 2 & 5 \\
		$n_Q$ & 10 & 15 & 5 & 10 & 15 & 5 \\
		$N$ & 8000 & 8000 & 6500 & 6500 & 8500 & 8000 \\
		\bottomrule
\end{tabular}
\end{sc}
\end{small}
\end{center}
\vskip -0.1in
\end{table*}

\appendix

\section{Evaluation Metrics}

For the sake of completeness, we formally define the the Zero-Shot Learning
(ZSL) and Generalized Zero-Shot Learning (GZSL) metrics we use. We evaluate
our framework with the \textit{average per-class} [top-1] \textit{accuracy}
for ZSL, defined as
\begin{equation}
    \label{eq:at1}
    \begin{split}
        acc_{\mathcal{Y}} = \frac{1}{\|\mathcal{Y}\|}
            \sum_{y\in\mathcal{Y}}\frac{\text{\# correct predictions in y}}
                {\text{\# samples in y}},
    \end{split}
\end{equation}
where in the case of ZSL $\mathcal{Y}$ are the unseen classes. For GZSL, we use
the \textit{harmonic mean} of the average per-class accuracy of seen classes
and that of unseen classes, defined as
\begin{equation}
    \label{eq:harmonic}
    \begin{split}
        H = 2 \frac{u\cdot s}{u + s},
    \end{split}
\end{equation}
where we define $u = acc_{\mathcal{Y}}$ for unseen classes and $s = acc_{\mathcal{Y}}$
for seen classes for convenience and adherence to established notation.

\section{Hyperparameters}

We present the rest of the hyperparameters for the pre-training of the \textbf{Few-Shot
Learning (FSL)} algorithm, the Prototypical Network (PN), in Table~\ref{tbl:hyperparameters_pretrain} and for the training within our framework,
\mbox{Z2FSL(f-VAEGAN, PN)},
in Table~\ref{tbl:hyperparameters}, both for \textbf{Zero-Shot Learning (ZSL)} and \textbf{Generalized Zero-shot Learning (GZSL)}.

\section{Classifier Fine-tuning}

After the joint training of the FSL classifier and the generator of the backbone, we
fine-tune the FSL classifier on samples generated by the generator conditioned on attributes
of the unseen classes. This basically means that the generator provides both the support and the query set.
We train for 25 episodes, using the same learning rate as in the other two
settings the classifier is trained. We also retain the hyperparameters of the episode in this training regime
the same as in the joint training of the FSL classifier and the generator. This process provides marginal
improvement, if any, and is generally inconsistent. Further work is required to stabilize it.

\section{Prototypical Network initialization}

We initialize the weight matrices of the PN by setting all the diagonal elements
equal to 1, while the rest are randomly sampled i.i.d. from
$\mathcal{N}(.\ ;0, 0.01)$. We do so to bias the PN to preserve its input space
structure as much as possible, which we expect to be somewhat discriminative
due to ResNet-101. Notice that we can do so because ResNet-101 yields non-negative
values and we use ReLU activations. This structure can be thought of as similar
to a residual layer \cite{he2016deep} routinely used in Convolutional Neural nets.
It is for this reason that all weight matrices in the PN are square. We observed an
increase in validation accuracy in all settings using this clever initialization
trick. We are unaware of another similar approach in the literature,
so further investigation may be warranted.

\section{Data Augmentation}

For the extra crops besides the original image, we crop the original image
starting from the desired corner and extending up to 80\% of each dimension
and finally resize the crop to match the original image's dimensions.

\end{document}